\pdfoutput=1
\documentclass[11pt]{article}

\usepackage{emnlp2021}

\usepackage{tabularx}
\newcolumntype{L}[1]{>{\raggedright\let\newline\\\arraybackslash\hspace{0pt}}m{#1}}

\usepackage{times}
\usepackage{latexsym}
\usepackage[T1]{fontenc}
\usepackage[utf8]{inputenc}
\usepackage{multirow}
\usepackage{microtype}
\usepackage{enumitem}
\usepackage{comment}
\usepackage{graphicx}

\usepackage{wrapfig}
\usepackage{arydshln}

\title{{\sc All} Dolphins Are Intelligent and {\sc Some} Are Friendly: \\ Probing 
BERT for Nouns' Semantic Properties 
and their Prototypicality}

\author{Marianna Apidianaki \\
  Department of Digital Humanities \\
  University of Helsinki \\
  Helsinki, Finland \\
  \texttt{marianna.apidianaki@helsinki.fi} \\\And
  Aina Gar\'i Soler \\
  Université Paris-Saclay \\
CNRS, LISN \\
91400, Orsay, France\\
  \texttt{aina.gari@lisn.fr} \\}

\begin{document}

\maketitle
\begin{abstract}
Large scale language models encode rich commonsense  
knowledge acquired through exposure to massive data during pre-training, but their understanding of 
entities and their  semantic properties 
is  unclear. 
We probe BERT \cite{devlin2019bert}  
for the properties of English nouns 
as expressed by adjectives
that do not restrict the reference scope of the noun they modify (as in \textit{red car}), 
but instead emphasise some inherent aspect (\textit{red strawberry}). 
We base our study on psycholinguistics datasets that capture the association strength between nouns and their semantic features.  
We probe BERT using cloze tasks and in a classification setting, and show that the model  has marginal knowledge of these features and their prevalence 
as expressed in these datasets. We discuss 
factors that make evaluation 
challenging  and impede drawing general conclusions about the models' knowledge of noun properties. 
Finally, we show that when tested in a fine-tuning setting addressing entailment, BERT 
successfully leverages the information needed for reasoning about the meaning of adjective-noun constructions outperforming previous methods.

\end{abstract}

\section{Introduction}

Adjectival modification is one of the main types of composition in natural language \cite{baroni-zamparelli-2010-nouns,guevara-2010-regression}. Adjectives in attributive position\footnote{Adjectives that appear immediately before the noun they modify and form part of the noun phrase (e.g., {\it white rabbit}), as opposed to adjectives in predicative position  that occur  after the noun (e.g., this {\it rabbit} is {\it white}).} usually have a restrictive role 
on the reference scope of the noun they modify, limiting the set of things it refers to (e.g., {\it white rabbits} $\sqsubset$ {\it rabbits}). 
This property of  adjectives   has interesting entailment implications, generally leading to adjective-noun (AN) constructions where the entailment relationship with the head noun holds (AN~$\models$~N) \cite{baroni-etal-2012-entailment}. Entailment is directional ({\it white rabbit} 
 $\models$ {\it rabbit} but {\it rabbit} 
 $\not\models$  {\it white rabbit}) \cite{kotlerman_etal2010}, unless modification is not restrictive. When A is prototypical of the N it modifies 
(as in {\it soft silk, red lobster, small blueberry}), its insertion does not reduce the scope of N or add new information, but rather emphasises   some inherent property \cite{pavlick-callison-burch-2016-babies}.  
In these cases, N and AN  denote the same set; they are in an equivalence relation ({\it red lobster} = {\it lobster})  and entailment is  symmetric.
\begin{table}[t]
    \begin{center}
\scalebox{0.8}{
\begin{tabular}{p{3.9cm}|p{4.8cm}}
    \multicolumn{2}{c}{\textbf{Masking Properties}} \\ \hline 
    {\sc singular} & a balloon is [MASK]. \\ \hline
    {\sc plural} & balloons are [MASK]. \\ \hline
    {\sc singular} + {\tt usually}  & a balloon is usually [MASK]. \\ 
    {\sc plural} + {\tt usually} & balloons are usually [MASK]. \\ \hline
   {\sc singular}+{\tt generally} & a balloon is generally [MASK]. \\ 
   {\sc plural} + {\tt generally} & balloons are generally [MASK]. \\ \hline
    {\sc singular} + {\tt can be} & a balloon can be [MASK]. \\ 
    {\sc plural} + {\tt can be} & balloons can be [MASK].  \\ \hline
    {\tt  most} + {\sc plural} & most balloons are [MASK]. \\ \hline
    {\tt all} + {\sc plural}  & all balloons are [MASK]. \\
    \hline 
    {\tt some} + {\sc plural} & some balloons are [MASK]. \\
    \hline \hline
    
    \multicolumn{2}{c}{{\rule{0pt}{2.7ex}\bf Masking Quantifiers}}\rule[-1ex]{0pt}{0pt} \\ \hline
    %\multicolumn{2}{c}{\rule{0pt}{2.7ex}{\bf Masking Quantifiers}}\rule[-1.2ex]{0pt}{0pt} \\ \hline
    \multicolumn{2}{l}{\hspace{-2mm}{\parbox{6cm}{\vspace{1.2mm} [MASK] balloons are colourful.}}} ({\sc all-most-some}) \\ 
    \multicolumn{2}{l}{{\hspace{-2mm}\parbox{6cm}{\vspace{1.2mm} [MASK] balloons are large.}}} ({\sc some-some-few}) \\ 
    \multicolumn{2}{l}{\hspace{-2mm}\parbox{6cm}{\vspace{1.2mm} [MASK] balloons are round.}} ({\sc most-some-no}) \\
    \hline 
    
    \end{tabular}
    }
    \end{center}
    \caption{Cloze statements for the noun {\it balloon} with  its properties \citep{mcraeetal-2005} and  quantifiers  masked. Parentheses in the lower part of the table contain the gold quantifiers in \citep{herbelot-vecchi-2015}.} 
    \label{tab:cloze}
\end{table}

The notion of prototypicality has a prominent place in the computational linguistics literature, mainly by reference to relationships between nouns \cite{roller-erk-2016-relations,vulic-etal-2017-hyperlex}. 
The prototypicality of adjectives has been understudied and is absent from  lexico-semantic resources such as WordNet \cite{Fellbaum1998} and HyperLex  \cite{vulic-etal-2017-hyperlex}. Alongside the theoretical interest of this linguistic property and its impact on the entailment properties of AN constructions, 
identifying prototypical adjectives has interesting practical implications. It can serve to retrieve information about the general concept ({\it silk, blueberry}) when queries include such AN pairs ({\it soft silk, small blueberry}), or to  discard adjectives that do not add new information about the noun they modify  in summarisation or sentence compression.

We investigate the knowledge that the BERT model  \cite{devlin2019bert} encodes about nouns' inherent properties as described in AN constructions. Although pre-trained language  models have been shown to encode rich factual  and commonsense knowledge \cite{petroni-etal-2019-language,Bouraoui2020}, little is known about their understanding of the  properties of the involved entities. We specifically explore whether BERT encodes information about the prototypical properties of the class of objects denoted by a noun 
(for example, that {\it all lobsters are red} and {\it blueberries are small}). 
We use a set of collected norms that describe important concept features \cite{mcraeetal-2005} and their associated quantifiers \cite{herbelot-vecchi-2015}.  We rely on  these data to derive cloze statements that we use to query BERT about noun properties, and to train BERT-based classifiers predicting these properties. We furthermore fine-tune BERT for entailment and test it in a task that involves AN constructions  \cite{pavlick-callison-burch-2016-babies}. Our cloze task results show that BERT has only marginal knowledge of noun properties and their prevalence, but can still successfully detect cases where the addition of an adjective does not alter the meaning of a sentence and where entailment is preserved.

\section{Related Work}

Compositionality in AN constructions  has been a central topic in distributional and formal semantics. \citet{journals/cogsci/MitchellL10} derive the meaning representation of a composite phrase from that of its constituents by performing algebraic operations (addition and multiplication) on distributional word semantic vectors, while  \citet{baroni-zamparelli-2010-nouns} and \citet{guevara-2010-regression}  derive composite vectors  through composition functions  learned from corpus-harvested phrase vectors. 
In our work, we represent AN phrases by combining the contextualised BERT representations of A and N in sentences where they occur, using algebraic operations. We also investigate the extent to which the representations of A and N in an AN phrase capture its meaning, since token-level BERT embeddings encode information from the surrounding context. 

We furthermore address the entailment relationship between N and ANs (N $\models$ AN). In the opposite direction (AN $\models$ N) entailment generally holds, i.e. almost all ANs entail their head noun ({\it red car} $\models$ {\it car}) \cite{baroni-etal-2012-entailment,kober-etal-2021-data}. Determining whether N $\models$ AN holds, however, depends on the semantic properties described by the adjective. We base this analysis on 
the AddOne dataset proposed by \citet{pavlick-callison-burch-2016-babies}. AddOne 
consists of sentence pairs that contain AN phrases annotated for entailment through crowdsourcing. This simplified entailment task differs from the classical RTE task \cite{daganetal2005} in that 
the premise and  hypothesis differ by only one atomic edit (insertion of A). 
We use this task as a proxy for prototypicality based on the assumption that adjectives describing typical properties of a noun do not modify its scope (e.g., {\it lobster} $\models$ {\it red lobster}, but {\it car} $\not\models$ {\it red car}). Prototypicality has been addressed in the literature mainly with respect to nouns, i.e. the typical hyponyms in a specific semantic class (e.g., {\it dog} $\models$ {\it animal}), or  member concepts that are most central to a  category \cite{roller-erk-2016-relations}. \citet{vulic-etal-2017-hyperlex} also address verb prototypicality in terms of  how typical of an action a verb is (e.g., ``Is TO RUN a type of TO MOVE?''). Our work extends this notion to adjectives describing noun properties in AN phrases. 

On the probing side, previous work explores the factual  and commonsense knowledge present in 
pretrained language models (LMs). The LAMA (LAnguage Model Analysis) probe proposed by  \citet{petroni-etal-2019-language} contains sets of facts from various knowledge sources.\footnote{Relations between entities stored in Wikidata, common sense relations between concepts from ConceptNet  \cite{speer-havasi-2012-representing}, and knowledge aimed for 
answering natural language questions in SQuAD \cite{rajpurkar-etal-2016-squad}.} Each fact is converted into a ``fill-in-the-blank'' cloze statement that is used to query the LM for a missing token. 
A model is considered to know a fact ([{\sc subject}, {\tt relation}, {\sc object}]  triple) if it can successfully predict masked tokens in cloze statements expressing this fact (e.g., {\sc Dante} {\tt was born} in~\underline{\hspace{7mm}}). The {\tt HasProperty} relation in LAMA (extracted from ConceptNet) is similar to our relation of interest as it links nouns to adjectives describing their properties. ConceptNet contains 
3,894 such pairs, 
but a close inspection of the data reveals several problematic cases (e.g., \textit{informal both}, \textit{divine forgive}, {\it ten 10}). 
Additionally, the cloze statements proposed for this dataset were automatically extracted from Open Mind Common Sense (OMCS)\footnote{\url{https://github.com/commonsense/omcs}} sentences and are often too long, including  irrelevant information that might confuse the model.\footnote{For example: ``To understand the event ``The monkey ate some bananas.'', it is important to know that Banana is [MASK]''. The ground truth adjective in this case is {\it yellow}.} \citet{jiang-etal-2020-know} demonstrate the impact of prompt quality
on LM probing but focus on relations involving encyclopedic knowledge (e.g., born/died in, profession, subclass).
\citet{Bouraoui2020} also explore the knowledge BERT has about lexical, morphological and commonsense relations (e.g., hypernymy, meronymy, plural,  cause-effect) through fine-tuning, 
but neither they address noun properties.

\section{Datasets}
\label{datasets}

{\bf \citet{mcraeetal-2005} dataset (MRD)}: Semantic feature norms are used in the field of psycholinguistics for studying human semantic representation and computation. 
MRD contains feature norms for 541 living 
and nonliving concepts collected from  725  participants in an annotation task. The annotators proposed  features they thought were important for each concept, covering physical (perceptual), functional and other properties. 
Among the collected 7,258 concept-feature pairs, we find that a dolphin is {\it intelligent, friendly}, and {\it lives in oceans}; and that a  chandelier is {\it hanging from ceilings} and {\it is  made of crystal}. The number of annotators who proposed each feature is also provided. The  dataset has been extensively used to investigate and improve the knowledge about object properties encoded by distributional models \citep{rubinstein-etal-2015-well}, word embeddings \citep{lucy-gauthier-2017-distributional, yang-etal-2018-extracting} and, more recently, contextual LMs \citep{forbes2019neural,hasegawa-etal-2020-word}. These studies do not focus on adjectival attributes but rather consider all proposed properties, 
or specific subsets such as visual properties. 
In our experiments, we explore noun properties through the ``{\sc is}\_{\sc adj}'' features of noun concepts present in MRD.  

\paragraph {\bf{\citet{herbelot-vecchi-2015} dataset (HVD)}}: HVD adds an extra level of quantification annotations to the  MRD norms. 
Three native speakers of English selected a natural language quantifier among [{\sc no, few, some, most, all}]\footnote{{\sc no} and {\sc few} labels were rarely used by the annotators and we consider them as describing cases of non typical attributes.} for each concept-feature ({\it C,f})  pair, expressing the ratio of {\it C} instances having feature {\it f} (e.g., {\sc all}  guitars are musical instruments, but {\sc some} guitars are electric). 
Quantification is 
important for semantic inference; it can serve to understand set relations (such as synonymy and hyponymy) and to derive logically entailed sentences.  
We use the HVD dataset to probe BERT for the prevalence of noun properties. 

\paragraph {\bf\citet{pavlick-callison-burch-2016-babies} dataset (Add\-One)}: The Addone dataset is focused on AN composition. It contains 5,560 sentence pairs involving an AN  pair ($s_N$, $s_{AN}$) which have been  manually annotated for entailment ($s_N$ $\models$ $s_{AN}$) by crowd workers. 
Addone sentence pairs differ by one atomic edit, the insertion 
of A ($s_N$: ``There are questions as to whether our \underline{culture} has changed.'',  $s_{AN}$: ``There are questions as to whether our \underline{traditional culture} has  changed.''). Sentences were collected from corpora of different genres and each pair was annotated with a score in a 5-point scale from 1 (contradiction) to 5 (entailment). Only the pairs with high agreement (same score assigned by 2 out of 3 annotators) were retained. 
We use the AddOne dataset to assess BERT's ability to detect entailment  in AN constructions.   
The dataset comes with a pre-defined split into training, development and test sets (83/10/7\%) which we use in our  experiments addressing entailment. 

\section{Cloze Task Experiments}

We use the {\tt bert-base-uncased} and  {\tt bert-large-uncased} models  pre-trained on the BookCorpus (Zhu et al., 2015) and on English Wikipedia \cite{devlin2019bert}. The models are trained using a cloze task where tokens of the input sequence are masked and the models learn to fill the slots, and a binary classification objective where they  need to predict whether a particular sentence follows a given sequence of words.

\subsection{Cloze Task Probing for Properties} \label{sec:clozeprobing}

We retrieve adjective modifiers of nouns in MRD found in the {\sc is}\_{\sc adj}  features describing a concept ({\it bouquet}: {\sc is}\_{\it colourful}; {\it panther}: {\sc is}\_{\it black}). There are  509 noun concepts with at least one {\sc is}\_{\sc adj} feature in MRD. We exclude  features involving multi-word attributes  ({\it coconut}: {\sc is{\it\_white\_inside}}, {\it raft}: {\sc is{\it \_tied\_together\_with\_rope}}) 
which we do not expect BERT to be able to predict.\footnote{We use “multi-word” to refer to attributes 
that involve multiple words separated with underscores. These are not necessarily idiomatic expressions.} 
The average number of features per noun is 3.12 (1,592 in total). Table  \ref{number_of_attributes_per_noun} shows the number of nouns having a specific number of features. We define a set of templates and generate 
cloze statements for each noun that serve as our queries to probe BERT for these attributes. 
We define templates using both the singular and plural forms of a noun, as shown in Table \ref{tab:cloze}.\footnote{We use the plural form of nouns given by the {\tt pattern} Python library and manually correct any errors.} We always use plural templates
for 26 nouns given in plural form in MRD,\footnote{{\it beans}, {\it beets}, {\it curtains}, {\it earmuffs}, {\it jeans}, {\it leotards}, {\it mittens}, {\it onions}, {\it pajamas}, {\it peas}, {\it scissors}, {\it skis}, {\it slippers}, {\it shelves}, {\it sandals}, {\it bolts}, {\it gloves}, {\it nylons}, {\it boots}, {\it screws}, {\it pants}, {\it tongs}, {\it trousers}, {\it drapes}, {\it pliers}, {\it socks}.} 
and  singular templates for mass and uncountable nouns.\footnote{{\it  rice}, {\it bread}, {\it football}.} 
We evaluate the quality of the predictions made by BERT for each slot by checking the presence of ground-truth (reference) 
MRD adjectives  
in the top-1, top-5 and top-10 
predictions ranked by probability. We compare BERT to a baseline that ranks by frequency all bigrams where a specific noun appears in 
the second position (``\underline{\hspace{5mm}}~\textit{bouquet}'') in Google Ngrams \cite{brants2006web},
excluding bigrams that contain stop words and punctuation.

\begin{table}[t]
    \centering
    \scalebox{0.85}{ 
    \begin{tabular}{p{2cm}|p{3mm}p{3mm}p{3mm}p{3mm}p{3mm}p{3mm}p{3mm}p{3mm}p{3mm}p{3mm}}
    \hline
    \# features & \parbox{3mm}{\centering 1} & \parbox{3mm}{\centering 2} & \parbox{3mm}{\centering 3} & \parbox{3mm}{\centering 4} & \parbox{3mm}{\centering 5} & \parbox{3mm}{\centering 6} & \parbox{3mm}{\centering 7} & \parbox{3mm}{\centering 8} & \parbox{3mm}{\centering 9} \\ \hline
    \# nouns & \parbox{3mm}{\centering 98} & \parbox{3mm}{\centering 124} & \parbox{3mm}{\centering 97} & \parbox{3mm}{\centering 76} & \parbox{3mm}{\centering 60} & \parbox{3mm}{\centering 35} & \parbox{3mm}{\centering 12} & \parbox{3mm}{\centering 6} & \parbox{3mm}{\centering 1}  \\ \hline
    \end{tabular}
    }
    \caption{Number of nouns with a specific number of {\sc is\_adj} features in MRD.} 
    \label{number_of_attributes_per_noun}
\end{table}

\begin{figure}[t!]
\parbox{7.8cm}{
  \includegraphics[width=\columnwidth]{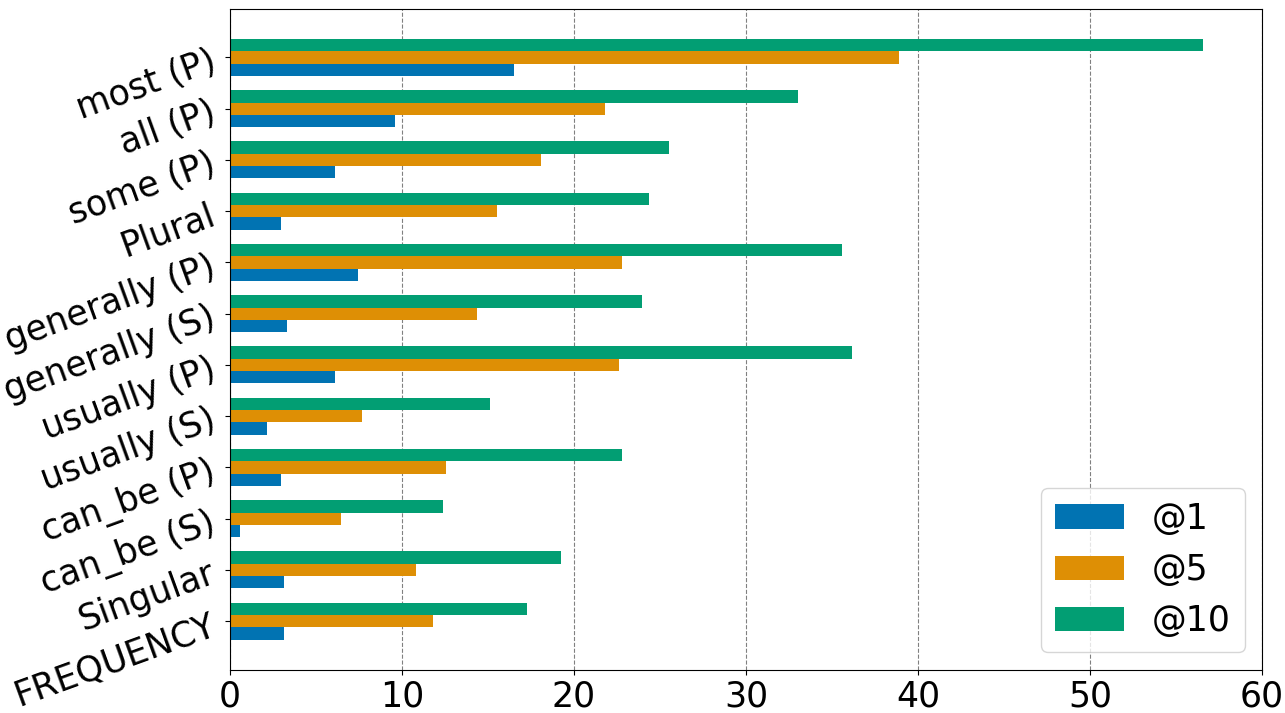}}
  \caption{Accuracy at top-{\it k} with  different  query templates  and  BERT-{\tt large}. 
  The (S) and (P) markers denote singular and plural templates. 
  Comparison to a frequency baseline.
  } 
  \label{number-words-rank} 
\end{figure}

Similar to  \citet{ettinger-2020-bert}, we define accuracy as the percentage of items (nouns) for which the expected completion (one of the reference 
properties) is among the model's top-{\it k} predictions. The plot in Figure \ref{number-words-rank} shows the accuracy of predictions at top-{\it k} for BERT-{\tt large}, with singular (S) and plural (P) queries. 
For {\it k} = 1, accuracy is very low, and naturally increases with {\it k} = 5 and {\it k}~=~10. We observe that results differ considerably when different cloze statements are used for probing. Overall, BERT makes less accurate predictions with 
singular templates (e.g., a balloon  usually is/can be   [MASK]) 
than  with plural templates (e.g., 
all/most balloons are [MASK]). Our assumption is that plural queries work better because they are more natural than singular 
ones,\footnote{For example, ``apples are red'' has a higher frequency (919) than ``an apple is red'' (278) in Google Ngrams.} and query naturalness seems to greatly influence prediction quality  \citep{ettinger-2020-bert}.
The frequency baseline 
proposes correct properties for more nouns than 
the {\sc singular} + {\tt usually} and {\sc singular} + {\tt can be}  
templates. 
BERT-{\tt large} gives slightly better results than BERT-{\tt base} on this task. Best results are obtained with 
{\tt most} + {\sc plural} queries (e.g., most balloons are [MASK]) where BERT-{\tt large} proposes a correct attribute for 287 out of 509 nouns (56.4\%).  
For BERT-{\tt base}, best results are retrieved with 
{\sc plural} + {\tt usually} 
queries (e.g., balloons are usually [MASK]), 
where a correct attribute is found in top-10 for 222 nouns. All results for BERT-{\tt base} are given in Appendix \ref{bertbase:results}.

The results of this probing experiment suggest that BERT has marginal knowledge of noun properties as reflected in the MRD association norms, and 
highlight the difficulty to retrieve this kind of information from the representations using cloze task probes. 
This information (e.g., ``bananas are yellow'') is rarely explicitly stated 
in texts, in contrast to other types of lexical and encyclopedic knowledge (e.g., hypernymy: ``a banana is a fruit'') available in the Wikipedia texts that were used for model pre-training. 
Another issue with a cloze  task evaluation for semantic properties is that contrary to structural properties (e.g., syntactic dependencies), there might be multiple correct answers for a query which might be partially covered by the resource used for evaluation. The quality of the resource, and the design of the task that served to gather the annotations, play an important role and should be taken into consideration for a fair interpretation of the results.  
We would, for example, expect annotators to propose a different set of properties if they were presented with cloze-type  queries. 

Although BERT does not always manage to predict the properties in MRD, we see from the quality of the proposed adjectives -- quite high in some cases -- that it encodes some knowledge about noun properties not present in the resource. 
For example, the  predictions made for the probe ``mittens are generally [MASK]'' might not contain the gold MRD adjectives ({\it knitted}, {\it colourful}),  but describe specific aspects such as 
their colour ({\it white, black, red, yellow}), 
shape and  composition ({\it flat, thick, short, thin}), 
and the fact that they can be {\it removed} (i.e. a garment). 
Naturally, the quality of the predictions varies a lot across nouns. 
These often describe 
general knowledge about the described entity, as is the case in the predictions made for the query ``all balloons are [MASK]'': 
\{{\it empty, free, flown, filled, lit, inflated, green, destroyed, closed, used}\}. 
We also observe that most completions proposed by BERT are adjectives (more details in Appendix \ref{app:proportion_adjectives}). 

BERT's predictions might contain synonyms of the adjectives present in MRD describing a correct property (for example, \textit{deadly} instead of  \textit{lethal} for the noun \textit{bomb}). We run an additional,  more relaxed, evaluation where we also consider as correct the adjectives' synonyms in WordNet \citep{Fellbaum1998}.\footnote{This is similar to synonym mapping in  
MT evaluation  \cite{banerjee-lavie-2005-meteor,marie-apidianaki-2015-alignment}.} As expected, we observe an increase in accuracy 
(cf. Appendix \ref{app:wordnet_eval}). This highlights the limitation of the string matching approach and the need for a more flexible evaluation methodology.

We also conduct a human evaluation of the predictions made by BERT-{\tt large} 
with the {\sc plural} + {\tt most} template for 90 nouns. 
Two subjects (post-graduates in linguistics) annotated the top ten predictions made by the model as correct or wrong (independent of whether they were present in MRD or another resource). The task is different than the one that served to create MRD in that the annotators were not asked to propose adjectives for a noun, but instead they had to judge whether a prediction described some property of the noun. The micro-average inter-annotator agreement as measured with the Cohen's kappa ($\kappa$) coefficient was fair (0.39)  \cite{LandisandKoch:1977,artstein-poesio-2008-survey}. This demonstrates that deciding whether an adjective describes a property of a noun (instead of a state or some other marginal feature) is difficult for human annotators.
We report the results of this evaluation in Appendix \ref{app:manual_eval}. 

Finally, we also investigate whether it is easier for BERT to propose correct properties for nouns 
that are not split into multiple tokens or WordPieces \citep{wu2016googles}. The results (cf. Appendix \ref{app:word_splitting}) confirm this intuition and show a slight decrease in the number of correctly predicted properties for nouns composed of multiple WordPieces.

\subsection{Cloze Task Probing for Quantifiers} \label{cloze_quant}

In order to probe BERT for prototypicality, we split the AN pairs 
in HVD into two sets. Set (A) contains 788 pairs (for 386 nouns)\footnote{Some nouns have several AN pairs each.} where the adjective describes a property that applies to most of the objects in the class denoted by the noun. These pairs are annotated with 
at least two {\sc all} labels, or with a combination of {\sc all} and {\sc most}  
(\textit{healthy banana} 
$\rightarrow$ [{\sc all-all-all]}).
We consider  adjectives 
in Set (A) as describing prototypical properties of the nouns.
Set (B), instead, 
contains 808 AN pairs (for 391 nouns) with adjectives describing properties of a smaller subset of the objects denoted by the noun. The  
 labels assigned to these pairs contain {\sc some}, {\sc few} and {\sc no}'s. 
For example, the annotations for \textit{red apple} are 
[{\sc most-some-some}], because there can be  green and yellow apples.\footnote{Note that a noun might be present in both Sets
(A) and (B), depending on whether its ANs describe prototypical properties. 
We find, for example, {\it transparent jar} 
in Set (A), because all jars have this property, and {\it breakable  jar} 
in Set (B), because not all jars can be easily broken.} 
We create 
cloze statements 
for the 788 pairs in Set (A), one for each pair, and for the 808 pairs in Set (B). A statement contains  the noun in plural form and a masked slot for the quantifier 
(for example, ``[MASK] {\it bananas are healthy}'', ``[MASK] {\it apples are red}''). We use each of the generated cloze statements to query BERT about  
prototypicality, checking whether it favours the expected (over the inappropriate) quantifier in its predictions. Given that BERT's predictions reflect the ranking of all words in the vocabulary according to whether they would be good fillers for the masked slot, we check the position of the quantifiers in the  
ranking.

\begin{table}[t]
    \begin{center}
\scalebox{0.8}{
\begin{tabular}{l||p{1.2cm}|p{1.8cm}||p{1.2cm}|p{1.8cm}} 
\multicolumn{1}{c}{} & \multicolumn{2}{c}{\parbox{1cm}{\centering {\bf Set A}}} & \multicolumn{2}{c}{\parbox{1cm}{\centering {\bf Set B}}}    \\  \hline
{\sc quant.}  & \parbox{1.2cm}{\centering MRR} & \parbox{1.9cm}{\centering \% of queries} & 
\parbox{1.2cm}{\centering MRR} & \parbox{1.9cm}{\centering \% of queries}  \\ \hline \hline
 \multicolumn{5}{c}{{\bf BERT-base}} \\ \hline 
{\sc  all} & \parbox{1.2cm}{\centering 0.203} & \parbox{1.8cm}{\centering 79.57} 
  & \parbox{1.2cm}{\centering 0.207} & \parbox{1.8cm}{\centering 75.25} 
  \\  \hline
{\sc  most} & \parbox{1.2cm}{\centering 0.167 }& \parbox{1.8cm}{\centering 64.47}
 & \parbox{1.2cm}{\centering 0.138} & \parbox{1.8cm}{\centering 55.07} 
 \\  \hline
 {\sc some} & \parbox{1.2cm}{\centering 0.188} & \parbox{1.8cm}{\centering 79.06}
  & \parbox{1.2cm}{\centering 0.156} & \parbox{1.8cm}{\centering 70.67} 
  \\ \hline 
  \multicolumn{5}{c}{{\bf BERT-large}} \\ \hline
  {\sc  all} & \parbox{1.2cm}{\centering 0.220} & \parbox{1.8cm}{\centering 75.13} 
  & \parbox{1.2cm}{\centering 0.221} & \parbox{1.8cm}{\centering 75.74} 
  \\  \hline
{\sc  most} &  \parbox{1.2cm}{\centering 0.235} & \parbox{1.8cm}{\centering 62.70} 
 & \parbox{1.2cm}{\centering 0.196} &   \parbox{1.8cm}{\centering 59.03}
  \\  \hline
 {\sc some} & \parbox{1.2cm}{\centering 0.201}  & \parbox{1.8cm}{\centering 69.54} 
  & \parbox{1.2cm}{\centering 0.166} &  \parbox{1.8cm}{\centering 65.34} 
  \\ 
  \hline
    \end{tabular}
    }
    \end{center}
    \caption{MRR and proportion of queries in Sets A and B where a quantifier is predicted in top-10.} 
    \label{tab:clozequantifiers}
\end{table}

We  evaluate the predictions using Mean Reciprocal Rank (MRR) (cf. Appendix \ref{app:mrr}). The  results are shown in Table \ref{tab:clozequantifiers}.  
The higher the MRR value is for a specific quantifier, 
the better its positionin the ranking. 
If BERT encoded the knowledge needed to distinguish prototypical from other properties, we would expect {\sc all} and {\sc most} to be 
higher in the ranking 
produced for Set (A) queries, and 
{\sc some} to come first in the ranking for Set (B) queries. We instead observe that {\sc all} tends to 
occupy a higher position in the predictions for both sets.  
In Table \ref{tab:clozequantifiers}, we also show the proportion of queries 
where a quantifier appears in the top-10 predictions.
We observe that all three quantifiers are often  proposed 
for queries in both sets. These results suggest that BERT 
does not distinguish properties on the basis of prototypicality. 

When several quantifiers are found in the top-10 predictions, 
we also check their relative position in the ranking. 
We calculate the percentage of queries where BERT assigned a higher probability to the expected quantifiers, 
ranking them higher than the others. 
This corresponds to the ``completion sensitivity test'' proposed by  \citet{ettinger-2020-bert},
which serves to explore BERT's ability to prefer expected over inappropriate completions. 
No clear precedence pattern is detected: BERT-{\tt base} assigns a higher probability to the expected ({\sc all}) than to the  incorrect completion ({\sc some}) in 56\% of Set (A) queries where both have been proposed; the inverse order is observed 
in 34\% of queries in Set (B). We also run the sensitivity test on the ranking obtained for the whole vocabulary,  
but no meaningful  patterns arise.\footnote{Appendix \ref{app:additional_quantifier_results} contains 
the quantifier precedence results.} 
The use of a sensitivity threshold \citep{ettinger-2020-bert}  turned out to be impractical in our setting because of the very low cloze probability assigned to the quantifiers in most cases.   
In our predictions, the definite article ``{\it the}'' is the most common top-1 prediction (in 85.7\% of  Set (A) queries, with an average probability of 0.629), followed by demonstrative and possessive determiners
(e.g., {\it their}, {\it these}). The probability mass is concentrated on these words, hence the probability assigned to quantifiers is often close to zero.\footnote{More details on determiners are given in Appendix \ref{app:additional_quantifier_results}.} 

We also check whether the observed trends reflect the prior probability of the quantifiers and of the definite article in a large corpus. 
We approximate this using their frequency in Google Ngrams.  
We find that $freq(\textsc{the}) > freq(\textsc{all}) > freq(\textsc{some}) > freq(\textsc{most})$. This is the same pattern obtained in our evaluation, with the exception of MRR results for 
BERT-{\tt large}, where {\sc most} is the highest ranked quantifier. 
This result suggests that BERT does not base prediction on the prevalence of noun properties but it, instead, largely follows the determiners' prior distribution.

\section{Classification Experiments}

We probe BERT representations for prototypicality 
in a classification setting, where models decide  
whether A describes a prototypical property of N. We use 
frozen embeddings (i.e. embeddings extracted from the pre-trained model)
and fine-tuning. 

\subsection{Experimental Setup}

\paragraph {\bf Examples} We consider as positive (prototypical)  instances ({\tt pos}) for this task AN phrases 
from HVD Set (A). 
As negative instances ({\tt neg}) for a noun in Set (A), we use the AN 
pairs where it appears in Set (B). 
If $|${\tt neg}$| < |${\tt pos}$|$ for a noun, 
we collect additional negative instances 
from the ukWaC corpus \citep{baroni2009wacky} 
where N is modified by an adjective A$^\prime$ 
such that A$^\prime$N $\notin$ HVD. We exclude cases where N is part of a compound (i.e. where it modifies another noun, as in \textit{small sardine tin}).\footnote{We obtain the dependency parse of a sentence using {\tt stanza} \cite{qi2020stanza}.} We retain the most frequent ANs 
found for N in ukWaC as negative instances, until $|${\tt neg}$| = |${\tt pos}$|$. 

Since common properties of nouns (e.g., {\it yellow banana}, {\it red strawberry}) are rarely explicitly stated in texts \cite{shwartz-choi-2020-neural}, 
we expect that the most frequent pairs found for a noun in ukWaC will not describe 
such properties. A manual exploration confirmed that the frequency-based filtering helps to retain good negative examples.  
The majority of the collected pairs do not describe prototypical properties (e.g., {\it useless pistol}, {\it organic celery}), with only a few ($\sim$10) exceptions 
(e.g., {\it silvery minnow}). The final dataset contains 1,566 instances in total, 783 for each class (positive and negative).\footnote{We omitted five positive AN pairs because not enough negative instances were found for the noun in Set (B) or in ukWaC.} 

\paragraph {\bf Representations} For each AN 
in  {\tt pos} and {\tt neg},
we obtain a BERT representation from a sentence ($s_{AN}$) in ukWaC where A modifies N. We pair $s_{AN}$ with a sentence $s_{N}$ where A has been automatically deleted (e.g., ``Then shape into balls about the size of a \underline{small tangerine}'' vs. ``Then shape into balls about the size of a \underline{tangerine}''). We choose sentences where A is not modified by an adverb (e.g.,  \textit{very small ant}, where removing \textit{small} would result in an ungrammatical sentence).
When no sentences are found for an AN 
(588 out of 1,566 cases), 
we use as $s_{AN}$ the plural pattern from the cloze task experiments (e.g.,  \textit{raspberries are edible}) 
and 
the plural noun alone as $s_N$  (\textit{raspberries}). When N is an uncountable noun, we use the singular pattern instead.\footnote{
Using sentences created with these patterns for all ANs hurts performance compared to the setting where sentences gathered from 
corpora are used.} We also feed 
the bigram in isolation into BERT (configuration that we call ISO). 

\paragraph{Data Split} We keep aside 10\% of the data as our development set and perform 5-fold cross-validation  on the rest. To minimise the impact of lexical memorisation where the  model learns that a word is representative of a specific class \citep{levy-etal-2015-supervised}, we observe a full lexical split by adjective between the development set and the data used for cross-validation, and also between the training and the test set in each fold. As a result, adjectives found in the test set 
at each iteration have not been seen in the training or in the development set. 
This is done to avoid that the model memorises an adjective as describing 
a common or prototypical property of nouns; 
for example, {\it small} is a feature for 120  out of 509 nouns in MRD. 
The split allows to evaluate the capability of the model to generalise to unseen adjectives.

\subsection{Embedding-based Classification} \label{sec:embbased_classif}

We expect the vector of an AN phrase involving a prototypical adjective ({\it red strawberry}) to be more similar to the vector of N 
({\it strawberry}), than that of a phrase A$^\prime$N 
involving an adjective that expresses a non typical property of N 
(\textit{rotten strawberry}).  
We extract three contextualised embeddings 
from each layer of the BERT-{\tt base} 
model  that we use  
to compare the representation of an AN phrase to that of the head N: 

\begin{enumerate}[noitemsep]
\item an embedding for N in sentence $s_N$ where N occurs without the adjective 
($\overrightarrow{Ns_{N}}$); 
\item an embedding for N in sentence $s_{AN}$ which contains the adjective ($\overrightarrow{Ns_{AN}}$);  
\item an embedding for A in $s_{AN}$ ($\overrightarrow{As_{AN}}$). 
\end{enumerate}

\noindent We obtain an AN 
representation ($\overrightarrow{AN}$) by combining the vectors pairwise: 
$\overrightarrow{Ns_{N}}$ and 
$\overrightarrow{Ns_{AN}}$; 
$\overrightarrow{Ns_{N}}$ and $\overrightarrow{As_{AN}}$; $\overrightarrow{Ns_{AN}}$ and 
$\overrightarrow{As_{AN}}$, 
using different composition operations: average, concatenation, difference, multiplication, and addition.
We also experiment with the 
token-level 
contextualised representations $\overrightarrow{As_{AN}}$ and $\overrightarrow{Ns_{AN}}$ only, which we expect to also encode information about the noun and the adjective in the AN, respectively, since they occur in the same context. 
We use the different AN representations as features for a logistic regression classifier. Additionally, we calculate the 
cosine similarity and  euclidean distance between the representation of a noun ($\overrightarrow{Ns_{N}}$ or 
$\overrightarrow{Ns_{AN}}$) and $\overrightarrow{AN}$ obtained through the  vector combinations and composition operations described above,  and feed them to the classifier as individual 
features or in combination. 
For comparison, we also run experiments using static word2vec \cite{mikolov2013efficient} and fastText 
\citep{mikolov-etal-2018-advances} embeddings as features, creating $\overrightarrow{AN}$ with the word embeddings $\overrightarrow{N}$ and $\overrightarrow{A}$, and using  $\overrightarrow{A}$ alone. For each type of representation (BERT, word2vec, fastText), we select the configuration with the highest average accuracy 
on the development set over the five cross-validation runs. In Table \ref{tab:embbased_results}, we report the average scores obtained on the test sets of the five folds for these configurations.  
Precision, recall and F1-score show how good a model is at detecting AN pairs that involve a prototypical adjective. As baselines, we provide results for a model that always predicts prototypicality ({\sc all-proto}), and a model that assigns the majority label found in the training set at each fold ({\sc majority}). 

\begin{table}[t]
    \centering
    \scalebox{0.9}{
    \begin{tabular}{l|cccc}
    \textbf{Model} & \textbf{Acc} & \textbf{F1} & \textbf{P} & \textbf{R}\\
    \hline
    BERT & \textbf{0.658} & 0.648 & \textbf{0.676} & 0.633 \\
    BERT (ISO) & 0.586 & 0.548 & 0.605 & 0.506 \\
    fastText & 0.593 & 0.481 & 0.639 & 0.411 \\
    word2vec & 0.559 & 0.455 & 0.601 & 0.372 \\
    \hline
    {\sc all-proto} & 0.507 & \textbf{0.672} & 0.507 & \textbf{1.000}\\
    {\sc majority} & 0.473 & 0.524 & 0.390 & 0.800 \\
    \end{tabular}}
    \caption{Average accuracy, F1-score, precision (P) and recall (R) of embedding-based classifiers on HVD in the cross-validation experiment across five folds.}
    \label{tab:embbased_results}
\end{table}

In terms of accuracy, BERT obtains the best results on this task (0.658) when 
cosine similarity and euclidean distance between 
$\overrightarrow{Ns_{N}}$ and  $\overrightarrow{Ns_{N}}$+$\overrightarrow{Ns_{AN}}$
at the last (12th) layer are used as features. 
The simple {\sc all-proto} baseline obtains the highest F1 score (0.672) but gets %a
low accuracy in this balanced dataset. 
Using sentences containing the AN is more effective than feeding 
the AN bigram in isolation into BERT (ISO). 
Static representations, especially word2vec, perform worse than BERT but still manage to beat the baselines in terms of accuracy. 
The best configuration for word2vec and fastText was the use of the 
adjective representations ($\overrightarrow{A}$) as features, which shows that the models do not manage to extract the 
information needed for assessing prototypicality from the different $\overrightarrow{N}$ and $\overrightarrow{A}$ combinations. 
Instead, the best strategy is to learn the tendency of an adjective to be prototypical.  When evaluated on unseen adjectives in our test sets, they base  prototypicality judgments on the similarity of these adjectives to the ones seen in the training set. 
We observe a high variation in accuracy and F1 scores across folds for all models. For BERT, F1 scores range from 0.553 to 0.740 and the range is even  larger for the fastText-based model (from 0.310 to 0.747).
This suggests that prototypicality is  not easy to detect for all AN pairs. 
Overall, BERT embeddings seem to be a better fit for estimating prototypicality than static representations. 
We report the detailed results by layer, and the best configurations per $\overrightarrow{AN}$ and composition type, in Appendix \ref{app:embedding_based_results}.

\begin{table}[]
    \centering
    \scalebox{0.9}{
    \begin{tabular}{l|cccc}
    \textbf{Model} & \textbf{Acc} & \textbf{F1} & \textbf{P} & \textbf{R}\\
    \hline
    BERT-CLS & 0.696 & \textbf{0.654} & 0.763 & \textbf{0.582} \\
    BERT-TOK & \textbf{0.697} & 0.646 & \textbf{0.778} & 0.561 \\
    \hline
    BERT-CLS (ISO) & 0.604 & 0.503 & 0.654 & 0.424 \\
    BERT-TOK (ISO) & 0.636 & 0.591 & 0.701 & 0.539 \\
    \end{tabular}}
    \caption{Average accuracy, F1 score, precision and recall in the cross-validation experiment across five folds for a BERT model fine-tuned on HVD using the CLS and TOK approaches.}
    \label{tab:ft_results}
\end{table}

\subsection{Fine-tuning BERT} \label{sec:mcrae_ft}

We compare our results with frozen 
embeddings to the performance of BERT fine-tuned for the prototypicality task. Specifically, we feed into BERT the two sentences in each ($s_N$, $s_{AN}$) pair separated by the 
[SEP] token. We experiment with a classifier on top of the  [CLS] token, as is typically done in sentence-pair classification tasks (we call this approach BERT-CLS); and with a classifier on top of the concatenation of two token representations: ($\overrightarrow{Ns_{N}}$, $\overrightarrow{As_{AN}}$), ($\overrightarrow{Ns_{N}}$, $\overrightarrow{Ns_{AN}}$), ($\overrightarrow{Ns_{N}}$, $\overrightarrow{As_{AN}} + \overrightarrow{Ns_{AN}}$) 
(our BERT-TOK approach). The two classification heads consist of a linear layer with softmax and are trained with a cross entropy loss. 
We fine-tune each model for 3 epochs with 0.1 dropout, and choose the learning rate based on the accuracy on the development set. 
Results of this experiment are found in Table \ref{tab:ft_results}. BERT-CLS and BERT-TOK ($\overrightarrow{Ns_{N}}$, $\overrightarrow{As_{AN}}$) perform comparably on this task and obtain better results than embedding-based models (Table \ref{tab:embbased_results}), with 0.697 accuracy. As in the experiment described in Section \ref{sec:embbased_classif}, using sentences yields better results than only feeding the AN (ISO).

\section{Entailment in AN Constructions}
\label{entailment-addone}
\subsection {Task Description}

AN constructions are often in a forward entailment relation with the head noun ({\it white rabbit}  $\models$ {\it rabbit})  \cite{baroni-etal-2012-entailment}. Whether backward entailment holds depends on the properties of N described by A. 
For example, a {\it car} is not always {\it red} (the label would be ``Unknown''), while {\it lobster} always entails {\it red lobster}. We explore BERT's capability to identify the AN cases where backward (N~$\models$~AN) entailment holds using the Addone dataset \cite{pavlick-callison-burch-2016-babies}. 

We fine-tune BERT on Addone to assess whether it captures the entailment relationship 
in AN constructions. BERT has shown high performance in other textual entailment tasks \cite{devlin2019bert}, 
but the Addone dataset has proved challenging for other models relying on recurrent 
architectures. 
We follow \citet{pavlick-callison-burch-2016-babies} and use Addone for a binary classification task, with the labels {\sc entailment} (for forward entailment and equivalence) and {\sc not entailment} (encompassing the contradiction, independence and reverse entailment relations). Similarly to the fine-tuning approach described in Section \ref{sec:mcrae_ft}, we feed into BERT 
the two sentences in each pair ($s_N$, $s_{AN}$) separated by the special [SEP] token. We again use the CLS and TOK classification heads. We fine-tune the model for 5 epochs with 0.1 dropout and select the learning rate based on the F1 score calculated over the actual {\sc entailment} cases 
  on the development set.\footnote{We use F1 score as a criterion, and not accuracy, because the Addone dataset is highly imbalanced (only 23\% of the instances belong to the {\sc entailment} class).}

\subsection{Results}

Results of our experiments on Addone  are presented in Table \ref{tab:addone_results}. We include results reported by \citet{pavlick-callison-burch-2016-babies} (P\&CB) for comparison. 
The {\sc maj} and {\sc maj-by-adj} baselines assign the majority class  in the training set ({\sc non-entailment}) and the majority class proposed for each adjective in the training set, respectively. We also report the human performance on this task as an upper bound, and compare to the best-performing model in \citet{pavlick-callison-burch-2016-babies} which relies on a RNN architecture \cite{bowman-etal-2015-large}. 
 BERT-CLS fails to learn the information needed for the task and predicts  the {\sc entailment} label for all instances. This explains the low  scores obtained with this model, since the majority label in this  dataset is {\sc non-entailment}. The default fine-tuning strategy used for textual entailment with BERT is, thus, not suitable for addressing cases of compositional entailment in the Addone dataset. It is much more effective to use the representations of the 
 specific words that determine sentence entailment: 
 BERT-TOK ($\overrightarrow{Ns_{N}}$, $\overrightarrow{As_{AN}}$) obtains higher results than the previous best model (RNN) and beats the {\sc maj} baseline in terms of accuracy, as well as 
 {\sc maj-by-adj} in terms of F1 
 and recall. 
 This shows that BERT leverages the   AN relations that are needed to solve this NLI task better compared to RNNs.

\begin{table}[t]
    \centering
    \scalebox{0.85}{
    \begin{tabular}{l|cccc}
         {\bf Model} & {\bf Acc} & {\bf F1} & {\bf P} & {\bf R} \\
         \hline
         Human {\sc (p\&cb)} & 0.933 & 0.730 & 0.840 & 0.640 \\ \hline\hline
         {\sc Maj-by-adj} {\sc (p\&cb)} & \textbf{0.922} & 0.680 & \textbf{0.860} & 0.560 \\
         {\sc Maj} {\sc (p\&cb)} & 0.853 & - & - & - \\
         \hline
         BERT-TOK & 0.912 & \textbf{0.696} & 0.709 & 0.684\\
         BERT-CLS & 0.147 & 0.257 & 0.147 & \textbf{1.000} \\
         \hline
         RNN {\sc (p\&cb)} & 0.873 & 0.510 & 0.600 & 0.440 \\
    \end{tabular}
    }
    \caption{Results on the Addone test set. Best results for each metric are highlighted in boldface.} 
    \label{tab:addone_results}
\end{table}

\section{Discussion}

Retrieving prototypical knowledge about entities is a real challenge for distributional models, not necessarily because of the models themselves and how advanced they are, but because this information is rarely stated in texts. 
This is described in the literature as the ``reporting bias'' phenomenon \cite{GordonandVanDurme:2013} which poses challenges on knowledge extraction. According to this phenomenon, rare actions or properties are over-represented in texts at the expense of trivial ones. 
\citet{shwartz-choi-2020-neural} show that the generalisation capability 
of pre-trained language models allows them to better estimate the plausibility of frequent but unspoken actions, outcomes and properties than previous models, but that they also tend to overestimate that of the very rare, amplifying the bias that already exists in their training corpus. In this study, using methodology commonly used for probing contextual 
models, we have precisely explored whether retrieving  knowledge about noun properties
constitutes a challenge for these models, or whether they manage to retrieve this knowledge due to their impressive generalisation capabilities.

 Since prototypical properties are often visual or perceptual, in future work we plan to combine text and visual features \citep{silberer-etal-2013-models,lazaridou-etal-2015-combining,Luetal:2019,li2019visualbert,li-etal-2021-unsupervised} for retrieving noun properties, in order to see how BERT can predict these properties 
 when it has access to images alongside text. Another goal is to collect evaluation data using cloze task queries in
specifically designed crowdsourcing tasks.
 
\section{Conclusion}

We have conducted 
a thorough investigation of the information encoded by  BERT  about nouns' intrinsic properties. 
Using datasets  specifically compiled for 
psycholinguistics studies, we probed 
BERT for noun properties and their prototypicality, as well as for the entailment relationship involving AN constructions  
which indicates possible generalisations to the entire class denoted by the noun. Our results show that information about noun properties, as described in 
word association norms, is hard to retrieve using cloze tasks. We discuss the limitations of 
semantic cloze tasks evaluation 
against  existing resources, and the need for more flexible evaluation scenarios. However,  
knowledge about properties can still be leveraged by BERT in a classification setting where the model is exposed to examples specifically encoding this information. 

We make our code and datasets available to promote further
research in this direction.\footnote{The code and datasets are available at
this URL:  \url{https://github.com/ainagari/prototypicality}}

\section*{Acknowledgements}
\setlength\intextsep{0mm}

\begin{wrapfigure}[]{l}{0pt}

\includegraphics[scale=0.3]{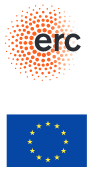}
\end{wrapfigure}

This work has been supported by the French National Research Agency under project ANR-16-CE33-0013. The work is also part of the FoTran project, funded by the European Research Council (ERC) under the European Union’s Horizon 2020 research and innovation programme (grant agreement \textnumero ~771113). We thank the anonymous reviewers for their helpful feedback and valuable suggestions.

\bibliography{anthology,custom}
\bibliographystyle{acl_natbib}

\appendix

\section{Masking Templates}

Table \ref{tab:templates} contains the templates that were used to construct the singular and plural queries for different nouns. SINGULAR\_NOUN and PLURAL\_NOUN are  placeholders for the nouns in  singular and plural form, respectively. 

\begin{table*}[t!]
    \begin{center}
\scalebox{0.8}{
\begin{tabular}{l|l}
     \hline 
    {\sc singular} & {\tt a SINGULAR\_NOUN is [MASK].} \\ \hline
    {\sc plural} & {\tt PLURAL\_NOUN are [MASK].} \\ \hline
    {\sc singular} + {\tt usually}  & {\tt a SINGULAR\_NOUN is usually [MASK].} \\ 
    {\sc plural} + {\tt usually} & {\tt PLURAL\_NOUN are usually [MASK].} \\ \hline
   {\sc singular} + {\tt generally} & {\tt a SINGULAR\_NOUN is generally [MASK].} \\ 
   {\sc plural} + {\tt generally} & {\tt PLURAL\_NOUN are generally [MASK].} \\ \hline
    {\sc singular} + {\tt can be} & {\tt a SINGULAR\_NOUN can be [MASK].} \\ 
    {\sc plural} + {\tt can be} & {\tt PLURAL\_NOUN can be [MASK].}  \\ \hline
    {\sc plural} + {\tt most} & {\tt most PLURAL\_NOUN are [MASK].} \\ \hline
    {\sc plural} + {\tt all}  & {\tt all PLURAL\_NOUN are [MASK].} \\
    \hline 
    {\sc plural} + {\tt some} & {\tt some PLURAL\_NOUN are [MASK].} \\
    \hline
    \end{tabular}
    }
    \end{center}
    \caption{Masking templates with  the noun in singular and plural form.} 
    \label{tab:templates}
\end{table*}

We generated the queries for the  quantifiers using the template: 

\parbox{7.4cm}{\vspace{2mm} \centering [MASK] {\tt PLURAL-NOUN are ADJECTIVE} .\vspace{2mm}} 

\noindent An example query generated using this template is: 

\parbox{7cm}{\vspace{2mm} \centering [MASK] balloons are colourful .\vspace{2mm}}

\section{Additional 
Masking Results} 

In this section of the Appendix, we present in more detail the results that we obtained in the masking experiments for noun properties and quantifiers. 

\subsection{Property Masking Results}
\label{bertbase:results}

\begin{figure}[t]
\centering
  \includegraphics[width=\columnwidth]{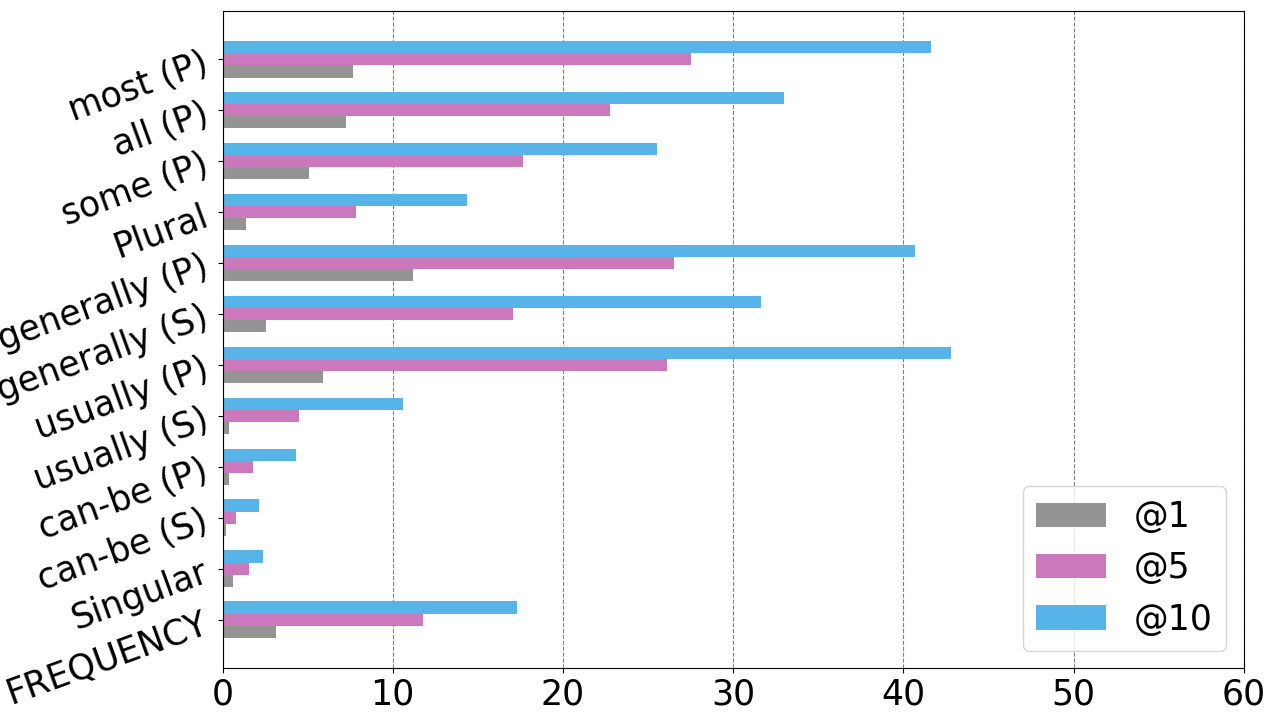}
  \caption{Accuracy at 
  top-{\it k} for  BERT-{\tt base}. (S) and (P) 
  stand for singular and plural templates. We compare to the results of a frequency baseline.
  } 
  \label{fig:number-words-rank} 
\end{figure}

\begin{figure}[] 
\begin{center}
  \includegraphics[width=\columnwidth]{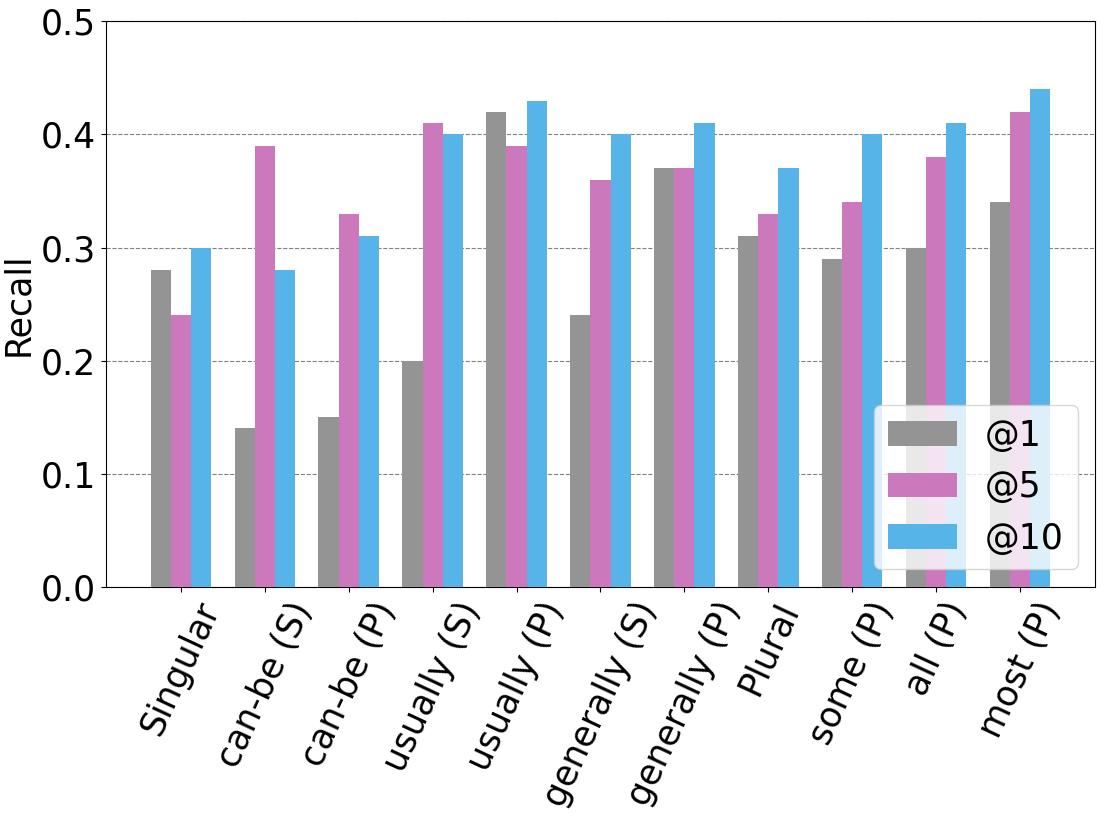}
  \caption{Average recall of MRD adjectives in the  top-$k$ predictions made by BERT--{\tt base}.} 
  \label{fig:recall-bertbase} 
    \end{center}
\end{figure}

\begin{figure}[] 
\begin{center}
  \includegraphics[width=\columnwidth]{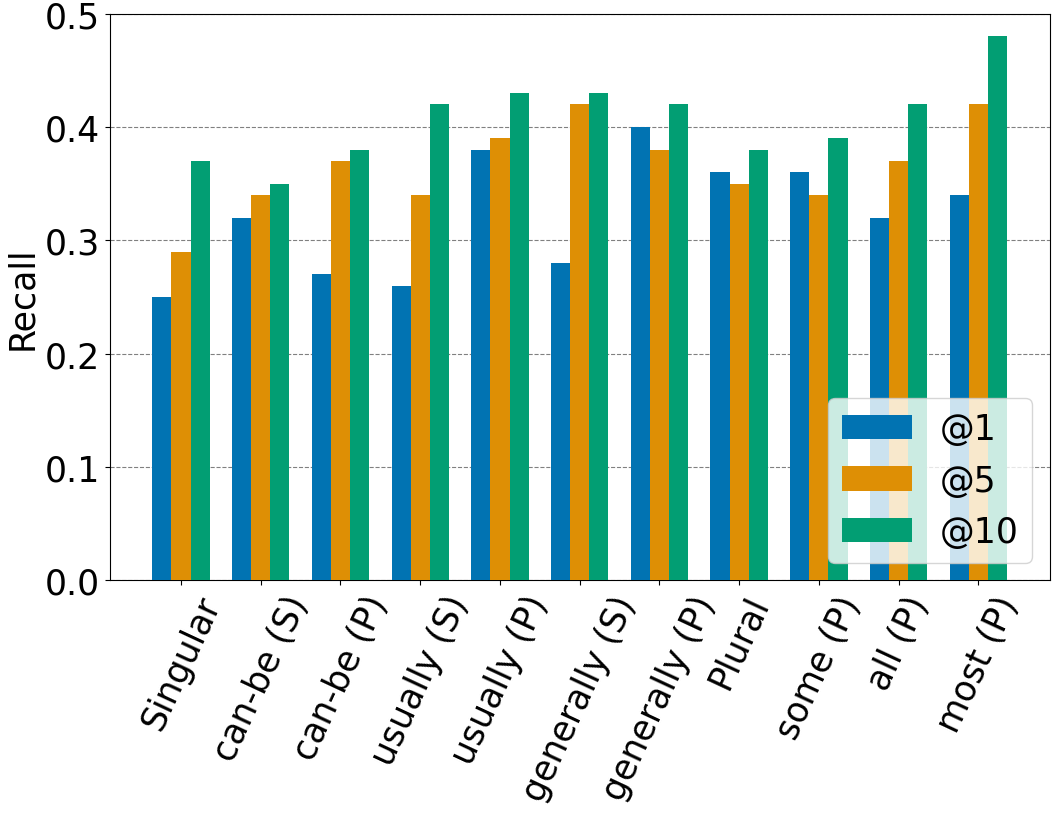}
  \caption{Average recall of MRD adjectives in the  top-$k$ predictions made by BERT--{\tt large}.} 
  \label{fig:recall-bertlarge} 
    \end{center}
\end{figure}

The plot in Figure \ref{fig:number-words-rank} shows the accuracy of predictions at top-\textit{k} for BERT-{\tt base}. Figures \ref{fig:recall-bertbase} and \ref{fig:recall-bertlarge} show 
 the average recall at the top-1, top-5 and top-10 positions 
 of the ranked BERT-{\tt base} and {\tt large} predictions, when using sentences constructed with the templates that correspond to the labels on the x axis. Average is calculated over the words for which at least one correct attribute is found at the specific rank.

\subsection{Adjectives in BERT Predictions} \label{app:proportion_adjectives}

\begin{figure}[t]
    \centering
    \includegraphics[width=\columnwidth]{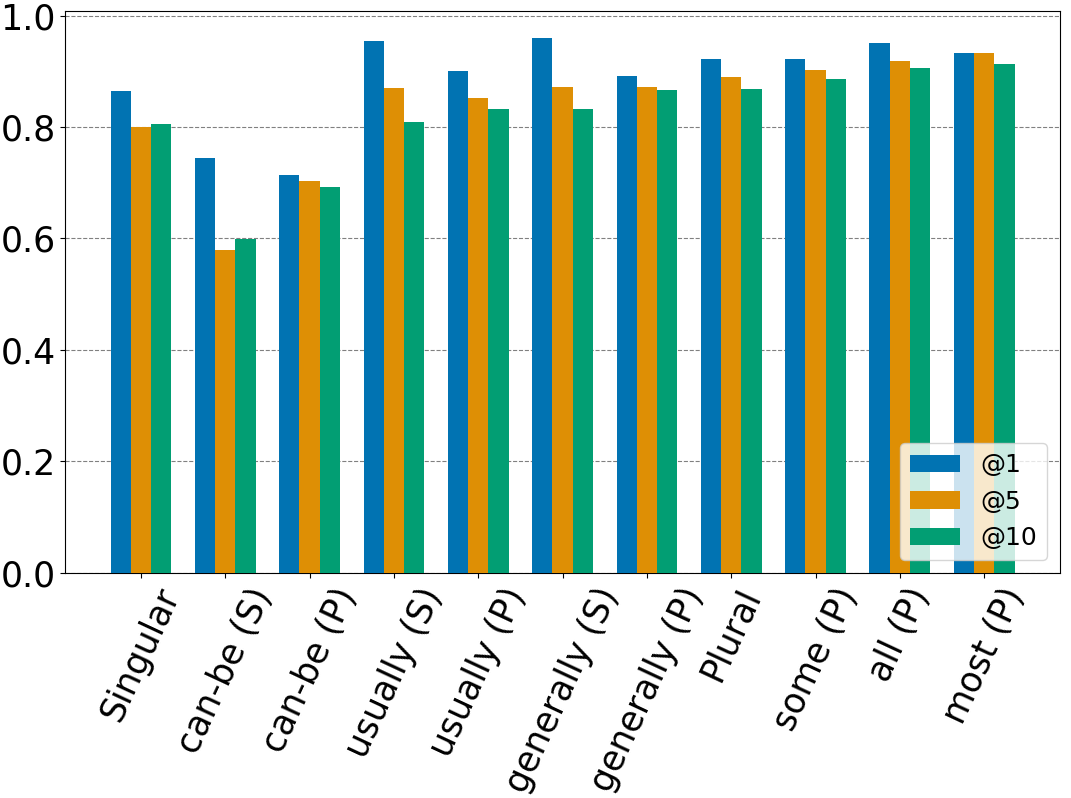} 
    \caption{Proportion of adjectives among BERT-{\tt large} predictions at different $k$.}
    \label{fig:proportion_adjectives_in_predictions}
\end{figure}

Figure \ref{fig:proportion_adjectives_in_predictions} shows the proportion of adjectives predicted by BERT-{\tt large} using different templates. We count as adjectives all words that have a synset of this part of speech in WordNet. We observe that the majority of predictions pertain to this part-of-speech. 
Fewer adjectives are proposed for 
queries 
of the form {\sc singular} + {\tt can be} (e.g., a balloon can be [MASK]), 
where BERT tends to favour 
verbs in the past participle form.

\subsection{WordNet-based Evaluation} \label{app:wordnet_eval}

\begin{figure}[t]
    \centering
    \includegraphics[width=\columnwidth]{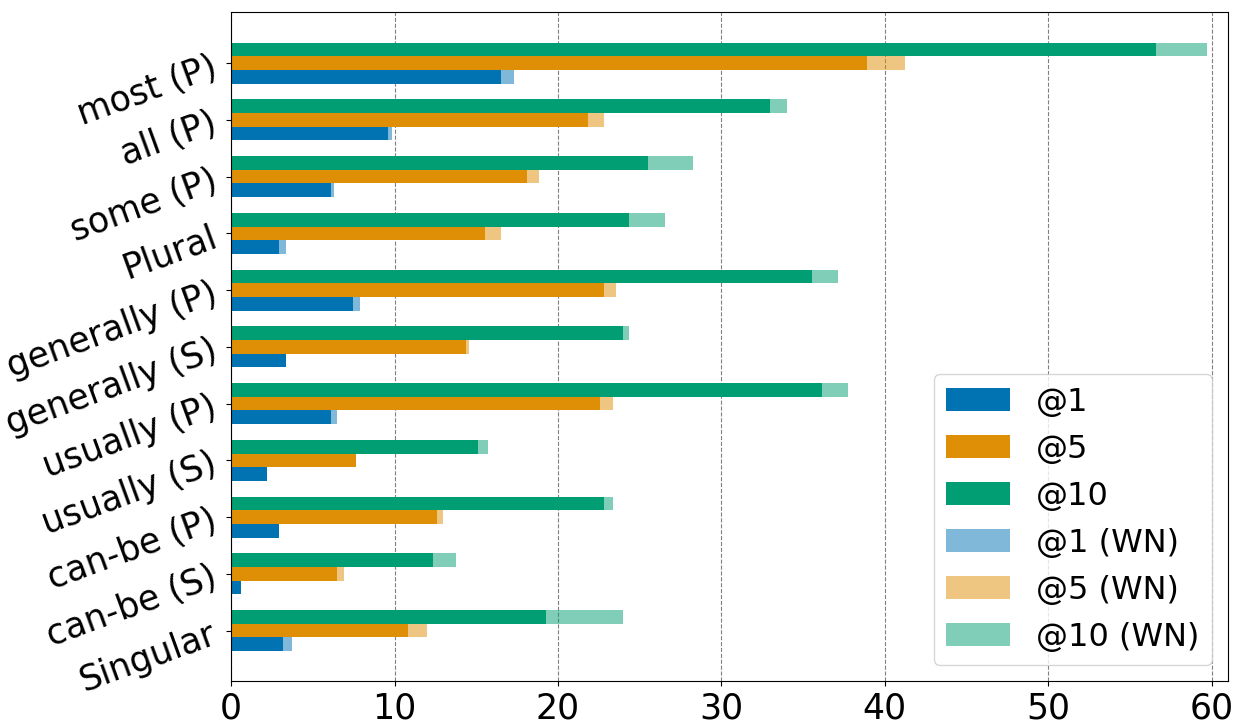} 
    \caption{Improvement in accuracy at top-$k$ for 
    BERT-{\tt large} predictions in 
    the WordNet-based (WN) evaluation.}
    \label{fig:wordnet_eval}
\end{figure}

Figure \ref{fig:wordnet_eval} presents the results of  our more relaxed evaluation, which includes 
the WordNet synonyms of the adjectives in MRD. Specifically, we expand the set of adjectives proposed in MRD for a noun (i.e. our reference) with their synonyms found in WordNet, and consider them all as correct. 
The lighter shades in the Figure show the improvement in accuracy at top-$k$ with respect to our previous results (darker shades). 
This shows that the model sometimes predicts correct properties that cannot be captured in an evaluation based solely on string matching. 

\subsection{Manual Evaluation of Predicted Properties} \label{app:manual_eval}

In Table  \ref{tab:manual_evaluation_proportions}, we present the results of the manual evaluation. We show the number of 
properties that were marked as correct by each annotator (A\#1 and A\#2). 
We also report the 
number of properties for  which the annotators agreed they were correct (Both). 
In the last column, we compare to the number of predictions that were found to be correct when evaluated solely against the reference properties in MRD.  
Since MRD has a different number of properties per noun -- often fewer than ten -- 
we indicate in parentheses the upper bound 
that could be reached if all reference properties for a noun were correctly predicted. 

Agreement between the annotators is fair (0.39), which demonstrates that deciding whether an adjective describes a property of a noun is difficult.
The annotators highlighted that there are some adjectives that BERT often proposes for nouns describing a specific class; for example, {\it nocturnal}, {\it solitary} and {\it shy} were proposed for different animals. We also find different colours proposed for {\it butterfly} ({\it white}, {\it black}, {\it brown}, {\it green}, {\it blue}) instead of the adjective {\it colourful} which describes a more general property of the insect (and which is one of its features in MRD).

\subsection{Impact of Word Splitting on Performance} \label{app:word_splitting}

BERT uses WordPiece tokenization \citep{wu2016googles}. The most frequent words are represented with a single token, but other words are 
split into multiple wordpieces. 
We investigate the 
impact of wordpiece splitting on the results. 
We run this analysis with BERT-{\tt large} predictions at top-10. 
We classify nouns into two classes, {\tt correct} and {\tt incorrect}, depending on whether at least one of their properties was correctly predicted.
We compare the proportion of multi-piece nouns (MPs)  in the two classes using $\chi^2$ tests.
We observe a significant difference ($\alpha=0.05$) 
with 4 out of our 11 templates. Table \ref{tab:chisquare} contains the p-values and effect sizes (Cramer's V) for these templates. We also report the proportion of MPs 
in the {\tt incorrect} class, which is slightly higher than that 
over all nouns in our dataset.\footnote{The number of 
MP nouns in singular and plural templates differs because plural forms of the nouns are more often split into multiple tokens.}  
The effect size values are also weak, suggesting 
that word splitting has only a slight negative effect on BERT's performance on this cloze task. 

\begin{table}[t]
\centering
\scalebox{0.9}{
\begin{tabular}{l||ccc|c}
    & \multicolumn{3}{l|}{\parbox{3.7cm}{\centering {\bf Manual evaluation} \vspace{1mm}}} & \\
    & {\bf A\#1}  & {\bf A\#2}  & {\bf Both}  & {\bf MRD}  \\
    \hline
@1  & 57        & 45      & 39  & 15 (90) \\
@5  & 228        & 166       & 130  & 50 (327) \\
@10 & 426      & 291       & 224  & 89 (355)      
\end{tabular}
}
\caption{Number of properties predicted by BERT-{\tt large} 
with the ``{\sc plural} + {\tt most}'' 
The last column shows the number of 
correct predictions when evaluating against properties in MRD. 
The upperbound for this evaluation is given in parentheses.} 
\label{tab:manual_evaluation_proportions}
\end{table}

\begin{table*}[]
\centering
\scalebox{0.9}{
\begin{tabular}{l||c|c|c|c}

                   & \textbf{p-value} & \begin{tabular}{@{}c@{}}\textbf{effect size}\\ \textbf{(Cramer's V)}\end{tabular} & \parbox{4cm}{\centering \textbf{\% of MPs with} \\ \textbf{no correct predictions}}
                   & \textbf{total \% MPs}
                    \\
                   \hline
{\sc singular} + {\tt usually} & 0.011   & 0.11 & 30.7\% &   \multirow{3}{*}{28.2\%}                                                           \\
{\sc singular} + {\tt generally} & 0.038   & 0.09 & 30.7\% &                                                             \\
{\sc singular} + {\tt can be} & 0.002   & 0.14  & 30.6\% &                                                             \\ \cline{5-5} 
%plural
{\sc plural} & 0.03    & 0.03 & 59.2\% & 56.4\%                                                           
\end{tabular}}
\caption{Statistics of the $\chi^2$ tests that showed a significant association between (a) a noun being multi-piece (MP), 
and (b) BERT predicting wrong properties for the noun (as evaluated against MRD). The last column shows the 
proportion of MP nouns in singular and plural  templates. 
}
\label{tab:chisquare}
\end{table*}
\begin{table}[th!]
    \begin{center}
\scalebox{0.83}{
\begin{tabular}{cccc}
& \multicolumn{2}{|c||}{\textbf{Results @10}} &
\textbf{Results @|V|} \\
\hline
\hline
 \multicolumn{4}{c}{{\bf BERT-base}} \\ \hline 
  {\sc quantifiers} & \multicolumn{1}{|c|}{{\bf \% 
  queries}} 
&  \begin{tabular}{@{}c@{}}{\bf \# 
queries}\end{tabular} 
   & \multicolumn{1}{||c}{{\bf \% 
   queries}} 
\\ \hline
 \multicolumn{4}{c}{{\bf Set (A)}} \\ \hline 
 {\sc all} $>$ {\sc  some} & 
 \multicolumn{1}{|c|}{56.02\%} &
 532 &
 \multicolumn{1}{||c}{53.43\%} \\
 {\sc most} $>$ {\sc some} & 
 \multicolumn{1}{|c|}{49.89\%} &
 451 & 
 \multicolumn{1}{||c}{37.31\%} \\
 \hline
 \multicolumn{4}{c}{{\bf Set (B)}} \\ \hline 
 {\sc some} $>$ {\sc all} & 
 \multicolumn{1}{|c|}{34.48\%} & \centering 467 & 
 \multicolumn{1}{||c}{40.1\%} \\
{\sc some} $>$ {\sc  most} & 
 \multicolumn{1}{|c|}{83.87\%} & \centering 31 & 
 \multicolumn{1}{||c}{65.47\%} \\
 \hline \hline
 \multicolumn{4}{c}{{\bf BERT-large}} \\ \hline 
 {\sc quantifiers} & \multicolumn{1}{|c|}{{\bf \% 
 queries}} 
&  \begin{tabular}{@{}c@{}}{\bf \# 
queries}\end{tabular} & \multicolumn{1}{||c}{{\bf \% 
queries}} \\ \hline 
 \multicolumn{4}{c}{{\bf Set (A)}} \\ \hline 
 {\sc all} $>$ {\sc  some} &  \multicolumn{1}{|c|}{55.19\%} & 462 & \multicolumn{1}{||c}{55.08\%} \\
 {\sc most} $>$ {\sc some} & \multicolumn{1}{|c|}{58.00\%} & 431 & \multicolumn{1}{||c}{44.16\%} \\
 \hline
 \multicolumn{4}{c}{{\bf Set (B)}} \\ \hline 
 {\sc some} $>$ {\sc all} &  \multicolumn{1}{|c|}{33.41\%} & 449 & \multicolumn{1}{||c}{35.02\%} \\
 {\sc some} $>$ {\sc  most} & \multicolumn{1}{|c|}{48.00\%} & 25 & \multicolumn{1}{||c}{51.98\%} \\
    \end{tabular}
    }
    \end{center}
    \caption{Relative position of correct vs. incorrect quantifiers when they both appear in the top-10 predictions made by each model, and in the ranking for the whole vocabulary.
}
    \label{tab:clozequantifiers_precedence}
\end{table}

\begin{table}[h!]
    \begin{center}
\scalebox{0.8}{
\begin{tabular}{p{0.7cm}|p{1.65cm}|p{1.7cm}||p{1.65cm}|p{1.7cm}} 
\multicolumn{5}{c}{\parbox{8cm}{\centering {\bf BERT-base}}} \\ \hline
& \multicolumn{2}{c}{\parbox{2cm}{\centering {\bf Set (A)}}} & \multicolumn{2}{c}{\parbox{2cm}{\centering {\bf Set (B)}}}    \\  \hline
& \parbox{1.7cm}{\centering {\% 
queries}} &  \parbox{1.7cm}{Avg. Prob.} & \parbox{1.65cm}{\centering {\% queries}} & \parbox{1.8cm}{Avg. Prob.}  \\  \hline
\parbox{0.8cm}{\centering {\it the}} & \parbox{1.7cm}{\centering 85.7\%} & \parbox{1.7cm}{\centering 0.629} & \parbox{2cm}{\centering 90.5\%} & \parbox{1.7cm}{\centering 0.662} \\  
\parbox{0.8cm}{\centering {\it these}} & \parbox{1.7cm}{\centering 6.2\%} & \parbox{1.7cm}{\centering 0.350} & \parbox{1.7cm}{\centering 3.8\%} & \parbox{1.7cm}{\centering 0.375} \\  
\parbox{0.8cm}{\centering {\it their}} & \parbox{1.7cm}{\centering 0.8\%} & \parbox{1.7cm}{\centering 0.523} & \parbox{2cm}{\centering 1.0\%} & \parbox{1.7cm}{\centering 0.686} \\  \hline \hline
\multicolumn{5}{c}{\parbox{8cm}{\centering {\bf BERT-large}}} \\ \hline
& \multicolumn{2}{c}{\parbox{2cm}{\centering {\bf Set (A)}}} & \multicolumn{2}{c}{\parbox{2cm}{\centering {\bf Set (B)}}}    \\  \hline
& \parbox{1.65cm}{\centering {\% queries}} &  \parbox{1.8cm}{Avg. Prob.} & \parbox{1.65cm}{\centering {\% queries}} & \parbox{1.8cm}{Avg. Prob.}  \\  \hline
\parbox{0.8cm}{\centering {\it the}} & \parbox{1.7cm}{\centering 74.3\%} & \parbox{1.7cm}{\centering 0.664} & \parbox{1.7cm}{\centering 79.9\%} & \parbox{1.7cm}{\centering 0.716} \\  
\parbox{0.8cm}{\centering {\it these}} & \parbox{1.7cm}{\centering 4.2\%} & \parbox{1.7cm}{\centering 0.419} & \parbox{1.7cm}{\centering 3.1\%} & \parbox{1.7cm}{\centering 0.510} \\  
\parbox{0.8cm}{\centering {\it their}} & \parbox{1.7cm}{\centering 0.4\%} & \parbox{1.7cm}{\centering 0.398} & \parbox{1.7cm}{\centering 0.7\%} & \parbox{1.7cm}{\centering 0.572} 
    \end{tabular}
    }
    \end{center}
    \caption{Proportion
    of queries in each set where the determiners {\it the}, {\it these} and {\it their} 
    are found at the first 
    position in the ranking. We also report the average probability assigned to a     determiner when found in this position.  
   }
    \label{tab:determiners}
\end{table}

\subsection{Additional Quantifier Probing Results} \label{app:additional_quantifier_results}

Table \ref{tab:clozequantifiers_precedence} shows the relative position 
of correct vs. incorrect quantifiers when they both appear in the top-10 predictions (columns 2 and 3), 
and in the ranking for the  whole vocabulary (column 4). 
Correct (expected) completions 
for Set (A) queries are {\sc all} and {\sc most}; for Set (B), 
the correct answer is {\sc some}. 

The symbol ``$>$'' in the  first column denotes precedence of a quantifier over another  (i.e. higher probability). Column 3 shows the number of queries for which the two quantifiers were proposed in top-10, which served to calculate the proportion in column 2. 
{\sc all} and {\sc some} were, for example, proposed by BERT-{\tt base} in top-10 for 532 Set (A) queries. {\sc all} had a higher probability than {\sc some} in 56\% of these queries.

Table \ref{tab:determiners} shows the ranking results for other determiners (\textit{the}, \textit{these}, \textit{their}) that are found in the first (top-1) position.

\section{Mean Reciprocal Ranking} \label{app:mrr}

Equation \ref{eq:mrr} contains the formula for Mean Reciprocal Ranking (MRR). RR measures the reciprocal of the rank  of the first correct answer, and MRR is the average across queries. We use MRR in Section \ref{cloze_quant} 
to evaluate quantifier prediction. $|Q|$ corresponds to the number of queries. 

\begin{equation}
\label{eq:mrr}
MRR = \frac{1}{|Q|}\sum_{i=1}^{|Q|}\frac{1}{rank_{i}}
\end{equation}

\vspace{2mm}
\begin{table}[t!]
    \centering
    \scalebox{0.95}{
    \begin{tabular}{lc||lc}
    \textbf{$\overrightarrow{AN}$ type}  & \textbf{Acc} & \textbf{composition} & \textbf{Acc}\\
    \hline
    $Ns_{N}$, $Ns_{AN}$ &  0.712 &  addition &  0.712 \\
    $As_{AN}$ &  0.675 &  difference & 0.667 \\
   $Ns_{N}$, $As_{AN}$ &  0.667  & concatenation &  0.660 \\
    $Ns_{AN}$, $As_{AN}$ & 0.665  & average & 0.650 \\
    $Ns_{AN}$ & 0.613 & multiplication & 0.611 \\
    \end{tabular}}
    \caption{Highest average accuracy obtained by the different types of AN representation (left) and composition operations (right) with BERT embedding-based classifiers on the HVD development set.}
    \label{tab:embbased_results_by_antype}
\end{table}

\begin{figure}[ht!]
    \centering
    \includegraphics[width=\columnwidth]{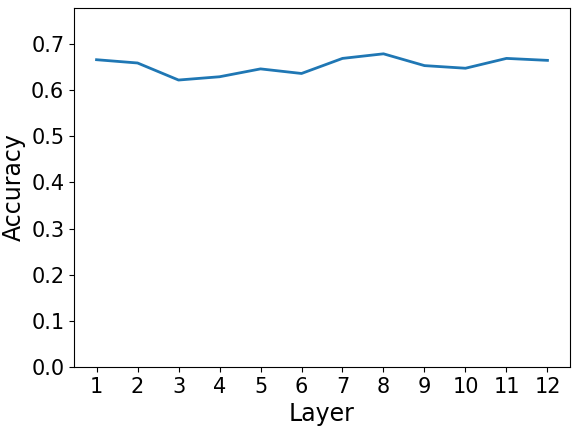}
    \caption{Highest average accuracy obtained by the embedding-based classifier on the HVD development set at every BERT layer.}
    \label{fig:bert_acc}
\end{figure}

\section{Detailed Embedding-based Classification results} \label{app:embedding_based_results}

In Table \ref{tab:detailed_dev}, we report the best results obtained on the HVD development set for each type of BERT-based $\overrightarrow{AN}$ representation and  composition operation. 
The combination of {$Ns_{N}$} and {$Ns_{AN}$}  
clearly outperforms the other vector combinations. 
Using the adjective token-level representation alone
($\overrightarrow{As_{AN}}$) 
also yields good results, 
definitely higher than 
$\overrightarrow{Ns_{AN}}$. In terms of composition functions, addition is the best performing 
operation for this task and multiplication the least useful.

Figure \ref{fig:bert_acc} shows the highest average accuracy obtained by each BERT layer on the HVD development set in these experiments.

\begin{table*}[]
    \centering
    \scalebox{0.90}{
    \begin{tabular}{lcccc}
    \hline
    \textbf{$\overrightarrow{AN}$ type}  & \textbf{Composition} & \textbf{Layer} & \textbf{Similarity} & \textbf{Accuracy}\\
    \hline
    $Ns_{N}$, $Ns_{AN}$ & addition & 12 & cosine \& euclidean ($Ns_{N}$, $Ns_{N}+Ns_{AN}$) & 0.712 \\
    $As_{N}$ & - & 8 & - & 0.675\\ 
    $Ns_{N}$, $As_{AN}$ & difference & 12 & - & 0.667\\
    $Ns_{AN}$, $As_{AN}$ & difference  & 12 & - & 0.665\\
    $Ns_{AN}$ & - & 11 & cosine ($Ns_{N}$, $Ns_{AN}$) & 0.613\\
    \hline
    \textbf{Composition type} & \parbox{2.5cm}{$\overrightarrow{AN}$ \textbf{type}}  & \textbf{Layer} & \textbf{Similarity}\\
    \hline
    addition & $Ns_{N}$, $Ns_{AN}$ & 12 & cosine \& euclidean ($Ns_{N}$, $Ns_{N}+Ns_{AN}$) & 0.712 \\
    difference & $Ns_{N}$, $As_{AN}$ & 12 & - & 0.667 \\
    concatenation & $Ns_{AN}$, $As_{AN}$ & 7 & - & 0.660 \\
    average & $Ns_{N}$, $As_{AN}$ & 5 & - & 0.650 \\
    multiplication & $Ns_{N}$, $As_{AN}$ & 7 & euclidean ($Ns_{N}$, $Ns_{N}\odot Ns_{AN}$) & 0.611\\
    \end{tabular}}
    \caption{Best configurations for each type 
    of $\overrightarrow{AN}$ (top) and composition operations (bottom) with BERT embedding-based  classifiers. These are identified based on the accuracy obtained on the HVD development set.}
    \label{tab:detailed_dev}
\end{table*}

\null
\vfill

\end{document}